\documentclass{article} 
\usepackage{iclr2023_conference, times}

\usepackage{amsmath,amsfonts,bm}

\def\eqref#1{equation~\ref{#1}}

\def\1{\bm{1}}

\DeclareMathAlphabet{\mathsfit}{\encodingdefault}{\sfdefault}{m}{sl}
\SetMathAlphabet{\mathsfit}{bold}{\encodingdefault}{\sfdefault}{bx}{n}

\usepackage{microtype}
\usepackage{graphicx}
\usepackage{subfigure}
\usepackage{booktabs} 

\usepackage{tabularx} 

\usepackage{hyperref}

\usepackage[shortlabels]{enumitem}

\usepackage{amsmath}
\usepackage{amssymb}
\usepackage{mathtools}
\usepackage{amsthm}

\usepackage[capitalize,noabbrev]{cleveref}

\theoremstyle{plain}

\theoremstyle{definition}

\theoremstyle{remark}

\usepackage[textsize=small]{todonotes}

\usepackage{silence}
\usepackage{colortbl}
\definecolor{Gray}{gray}{0.60}

\newcommand{\caffil}[1]{\textsuperscript{\mdseries #1}}
\usepackage{url}

\title{Med-Flamingo: a Multimodal Medical Few-shot Learner}

\author{Michael Moor\thanks{These authors contributed equally to this work.}\, \caffil{1} \ Qian Huang\footnotemark[1]\, \caffil{1} \ Shirley Wu\caffil{1} \ Michihiro Yasunaga\caffil{1} \ Cyril Zakka\caffil{2} \\ 
\textbf{Yash Dalmia\caffil{1} \ Eduardo Pontes Reis\caffil{3} \ Pranav Rajpurkar\caffil{4} \ Jure Leskovec\caffil{1}}  \\
}

\iclrfinalcopy 

\begin{document}

\maketitle
\vspace{-1cm}
\begin{center}
\textsuperscript{1}Department of Computer Science, Stanford University, Stanford,  USA \\
\textsuperscript{2}Department of Cardiothoracic Surgery, Stanford Medicine, Stanford,  USA \\
\textsuperscript{3}Hospital Israelita Albert Einstein, São Paulo, Brazil \\
\textsuperscript{4}Department of Biomedical Informatics, Harvard Medical School, Boston, USA \\
\end{center}

\begin{abstract}
Medicine, by its nature, is a multifaceted domain that requires the synthesis of information across various modalities. 
Medical generative vision-language models~(VLMs) make a first step in this direction and promise many exciting clinical applications. However, existing models typically have to be fine-tuned on sizeable down-stream datasets, which poses a significant limitation as in many medical applications data is scarce, necessitating models that are capable of learning from few examples in real-time.
Here we propose Med-Flamingo, a multimodal few-shot learner adapted to the medical domain. Based on OpenFlamingo-9B, we continue pre-training on paired and interleaved medical image-text data from publications and textbooks. Med-Flamingo unlocks few-shot generative medical visual question answering~(VQA) abilities, which we evaluate on several datasets including a novel challenging open-ended VQA dataset of visual USMLE-style problems. Furthermore, we conduct the first human evaluation for generative medical VQA where physicians review the problems and blinded generations in an interactive app. Med-Flamingo improves performance in generative medical VQA by up to 20\% in clinician's rating and firstly enables multimodal medical few-shot adaptations, such as rationale generation. We release our model, code, and evaluation app under~\url{https://github.com/snap-stanford/med-flamingo}.

\end{abstract}

\section{Introduction}

Large, pre-trained models (or foundation models)~have demonstrated remarkable capabilities in solving an abundance of tasks by being provided only a few labeled examples as context~\cite{bommasani2021opportunities}. This is known as in-context learning \cite{brown2020},  through which a model learns a task from a few provided examples specifically during prompting and without tuning the model parameters. In the medical domain, this bears great potential to vastly expand the capabilities of existing medical AI models~\cite{moor2023foundation}.  
Most notably, it will enable medical AI models to handle the various rare cases faced by clinicians every day in a unified way, to provide relevant rationales to justify their statements, and to easily customize model generations to specific use cases.

Implementing the in-context learning capability in a medical setting is challenging due to the inherent complexity and multimodality of medical data and the diversity of tasks to be solved. 

Previous efforts to create multimodal medical foundation models, such as ChexZero~\cite{tiu_expert-level_2022} and BiomedCLIP~\cite{zhang2023large}, have made significant strides in their respective domains. ChexZero specializes in chest X-ray interpretation, while BiomedCLIP has been trained on more diverse images paired with captions from the biomedical literature. Other models have also been developed for electronic health record~(EHR) data~\cite{steinberg2021language} and surgical videos~\cite{kiyasseh2023vision}. However, none of these models have embraced in-context learning for the multimodal medical domain. 
Existing medical VLMs, such as MedVINT~\cite{zhang2023pmc}, are typically trained on paired image-text data with a single image in the context, as opposed to more general streams of text that are interleaved with multiple images. Therefore, these models were not designed and tested to perform multimodal in-context learning with few-shot examples\footnote{For example, a challenge with multimodal in-context learning for existing medical vision language models is the potential for image information to leak across examples, potentially misleading the model.} 

Here, we propose Med-Flamingo, the first medical foundation model that can perform multimodal in-context learning specialized for the medical domain. Med-Flamingo is a vision-language model based on Flamingo~\citep{alayrac2022flamingo} that can naturally ingest data with interleaved modalities~(images and text), to generate text conditioned on this multimodal input. Building on the success of Flamingo, which was among the first vision-language models to exhibit in-context learning and few-shot learning abilities, Med-Flamingo extends these capabilities to the medical domain by pre-training on multimodal knowledge sources across medical disciplines. 
In preparation for the training of Med-Flamingo, our initial step involved constructing a unique, interleaved image-text dataset, which was derived from an extensive collection of over $4K$ medical textbooks (Section \ref{sec:pretrain}). Given the critical nature of accuracy and precision within the medical field, it is important to note that the quality, reliability, and source of the training data can considerably shape the results. Therefore, to ensure accuracy in medical facts, we meticulously curated our dataset from respected and authoritative sources of medical knowledge, as opposed to relying on potentially unreliable web-sourced data. 

\begin{figure*}[tbp]
\centering
\includegraphics[width=1\linewidth]{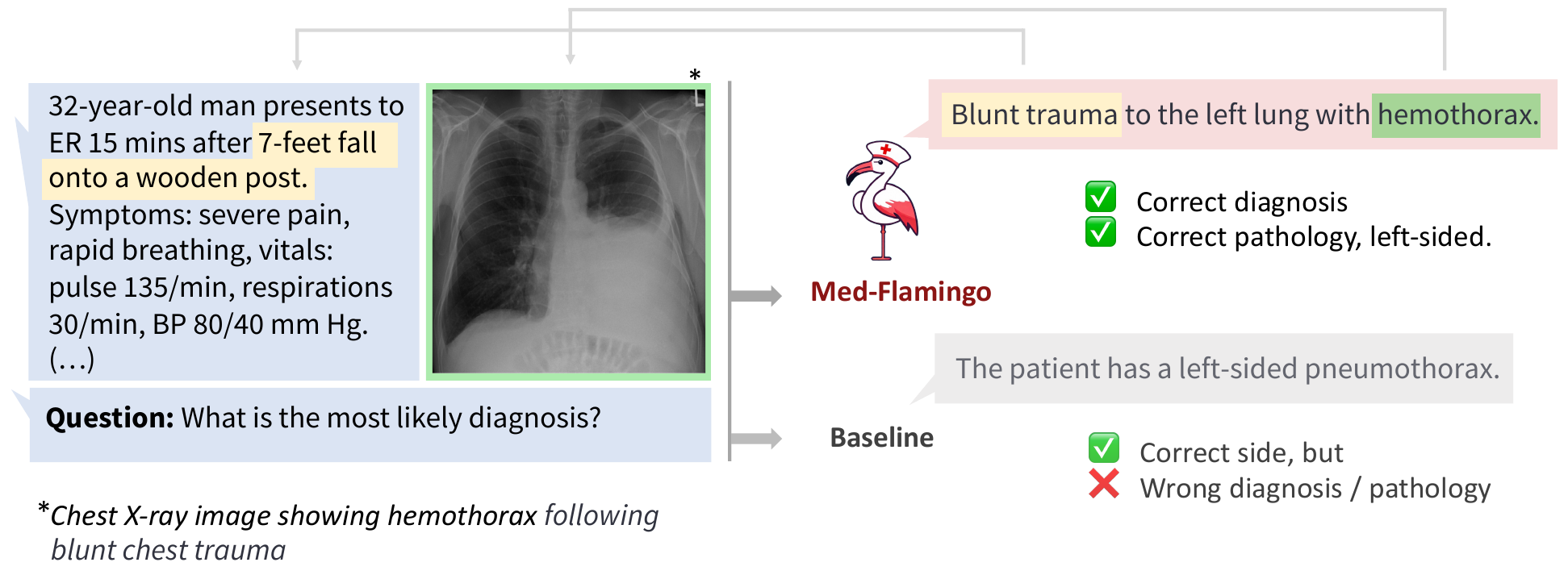}
\caption{Example of how Med-Flamingo  answers complex multimodal medical questions by generating open-ended responses conditioned on textual and visual information.
}
\label{fig:teaser}
\end{figure*}

In our experiments, we evaluate Med-Flamingo on generative medical visual question-answering~(VQA) tasks by directly generating open-ended answers, as opposed to scoring artificial answer options {\em ex post}--as CLIP-based medical vision-language models do. We design a new realistic evaluation protocol to measure the model generations' clinical usefulness. For this, we conduct an in-depth human evaluation study with clinical experts which results in a human evaluation score that serves as our main metric. In addition, due to existing medical VQA datasets being narrowly focused on image interpretation among the specialties of radiology and pathology, we create Visual USMLE, a challenging generative VQA dataset of complex USMLE-style problems across specialties, which are augmented with images, case vignettes, and potentially with lab results.

\begin{figure}
\centering

\includegraphics[width=1\linewidth]{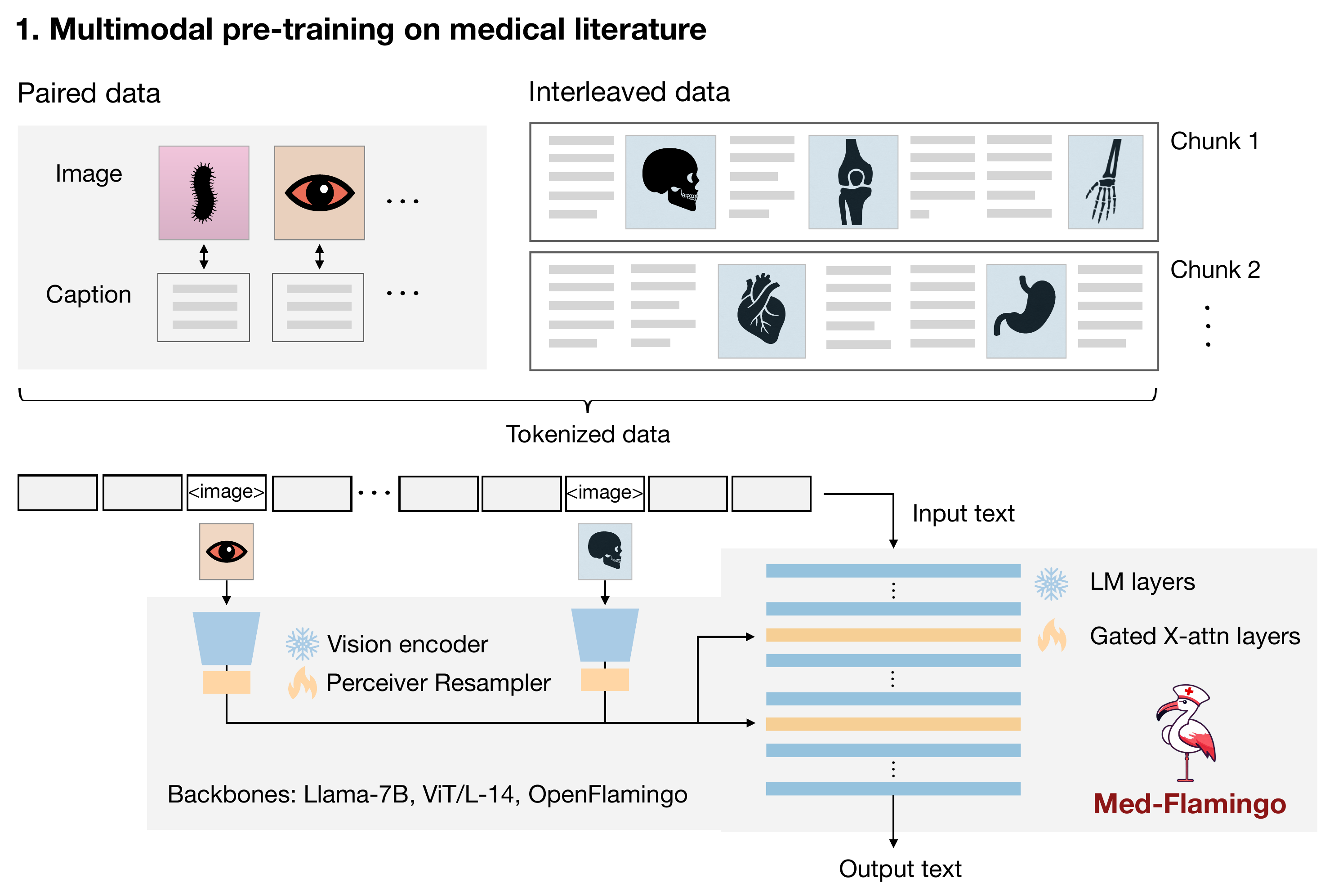} 
\vskip0.5cm
\includegraphics[width=1\linewidth]{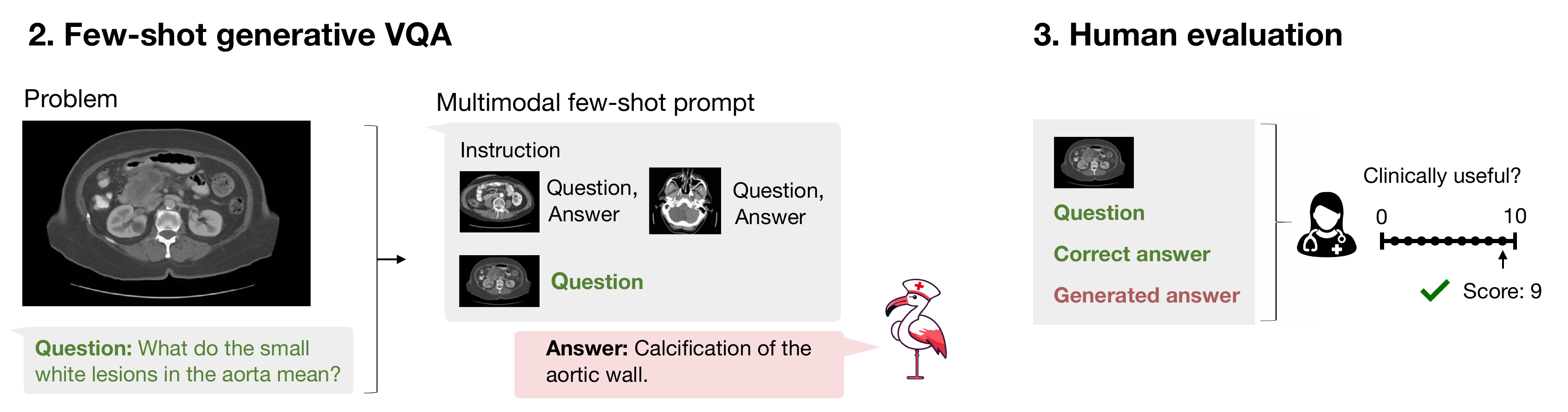}
\caption{Overview of the Med-Flamingo model and the three steps of our study. First, we pre-train our Med-Flamingo model using paired and interleaved image-text data from the general medical domain~(sourced from publications and textbooks). We initialize our model at the OpenFlamingo checkpoint continue pre-training on medical image-text data. Second, we perform few-shot generative visual question answering~(VQA). For this, we leverage two existing medical VQA datasets, and a new one, Visual USMLE. Third, we conduct a human rater study with clinicians to rate generations in the context of a given image, question and correct answer. The human evaluation was conducted with a dedicated app and results in a clinical evaluation score that serves as our main metric for evaluation.}
\end{figure}

Averaged across three generative medical VQA datasets, few-shot prompted Med-Flamingo achieves the best average rank in clinical evaluation score~(rank of $1.67$, best prior model has $2.33$), indicating that the model generates answers that are most preferred by clinicians, with up to 20\% improvement over prior models. 
Furthermore, Med-Flamingo is capable of performing medical reasoning, such as answering complex medical questions~(such as visually grounded USMLE-style questions) and providing explanations~(i.e., rationales), a capability not previously demonstrated by other multimodal medical foundation models. However, it is important to note that Med-Flamingo's performance may be limited by the availability and diversity of training data, as well as the complexity of certain medical tasks. All investigated models and baselines would occasionally hallucinate or generate low-quality responses. Despite these limitations, our work represents a significant step forward in the development of multimodal medical foundation models and their ability to perform multimodal in-context learning in the medical domain. We release the Med-Flamingo-9B checkpoint for further research, and make our code available under~\url{https://github.com/snap-stanford/med-flamingo}. 
In summary, our paper makes the following contributions:
\begin{enumerate} 
\item We present the first multimodal few-shot learner adapted to the medical domain, which promises novel clinical applications such as rationale generation and conditioning on retrieved multimodal context.
\item We create a novel dataset that enables the pre-training of a multimodal few-shot learner for the general medical domain.

\item We create a novel USMLE-style evaluation dataset that combines medical VQA with complex, across-specialty medical reasoning. 

\item We highlight shortcomings of existing evaluation strategies, and conduct an in-depth clinical evaluation study of open-ended VQA generations with medical raters using a dedicated evaluation app.
\end{enumerate}

\section{Related works}
The success of large language models (LLMs) \cite{brown_language_2020, liang2022holistic, qin2023chatgpt} has led to significant advancements in training specialized models for the medical domain. This has resulted in the emergence of various models, including BioBERT \cite{lee2020biobert}, ClinicalBERT \cite{huang2019clinicalbert}, PubMedBERT \cite{gu2021domain}, BioLinkBERT \cite{yasunaga_linkbert_2022}, DRAGON \cite{yasunaga_deep_2022}, BioMedLM \cite{biomedlm}, BioGPT \cite{luo2022biogpt}, and Med-PaLM \cite{singhal_large_2022}. Although these medical language models are typically smaller than general-purpose LLMs like GPT-3 \cite{brown_language_2020}, they can match or even surpass their performance on medical tasks, such as medical question answering.

Recently, there has been a growing interest in extending language models to handle vision-language multimodal data and tasks \cite{su2019vl, ramesh_zero-shot_2021, alayrac2022flamingo, aghajanyan_cm3_2022, yasunaga2022retrieval}. Furthermore, many medical applications involve multimodal information, such as radiology tasks that require the analysis of both X-ray images and radiology reports~\cite{tiu_expert-level_2022}. Motivated by these factors, we present a medical vision-language model (VLM). Existing medical VLMs include BiomedCLIP \cite{zhang2023large}, MedVINT~\cite{zhang2023pmc}. While BiomedCLIP is an encoder-only model, our focus lies in developing a generative VLM, demonstrating superior performance compared to MedVINT. Finally, Llava-Med is another recent medical generative VLM~\cite{li2023llava}, however the model was not yet available for benchmarking.

\begin{figure*}[h]
\centering
\includegraphics[width=1\linewidth]{figures/books_image.pdf}
\caption{Overview of the distribution of medical textbook categories of the MTB dataset. We classify each book title into one of the 49 manually created categories or "other" using the Claude-1 model.}
\label{fig:books}
\end{figure*}
\section{Med-Flamingo} \label{sec:pretrain}

To train a Flamingo model adapted to the medical domain, we leverage the pre-trained OpenFlamingo-9B model checkpoint~\cite{awadalla2023}, which is a general-domain VLM that was built on top of the frozen language model LLaMA-7B~\cite{touvron2023llama} and frozen vision encoder CLIP ViT/L-14~\cite{radford_learning_2021}. We perform continued pre-training in the medical domain which results in the model we refer to as Med-Flamingo.

\subsection{Data}
We pre-train Med-Flamingo by jointly training on interleaved image-text data and paired image-text data. As for the interleaved dataset, we created a interleaved dataset from a set of medical textbooks, which we subsequently refer to as \textsc{MTB}. As for the paired datasets, we used \textsc{PMC-OA}~\cite{pmcclip}. 

\paragraph{MTB} We construct a new multimodal dataset from a set of $4\,721$ textbooks from different medical specialties~(see Figure~\ref{fig:books}). During preprocessing, each book is first converted from PDF to HTML with all tags removed, except the image tags are converted to $<$image$>$ tokens. We then carry out data cleaning via deduplication and content filtering. Finally, each book with cleaned text and images is then chopped into segments for pretraining so that each segment contains at least one image and up to 10 images and a maximum length.  In total, MTB consists of approximately 0.8M images and 584M tokens. We use  95\% of the data for training and 5\% of the data for evaluation during the pre-training.

\paragraph{PMC-OA} 
We adopt the PMC-OA dataset~\cite{pmcclip} which is a biomedical dataset with 1.6M image-caption
pairs collected from PubMedCentral’s OpenAccess subset. We use 1.3M image-caption pairs for training and 0.16M pairs for evaluation following the public split\footnote{\url{https://huggingface.co/datasets/axiong/pmc_oa_beta}}.

\subsection{Objectives}
We follow the original Flamingo model approach~\cite{alayrac_flamingo_2022}, which considers the following language modelling problem: 

\begin{align*}
p\left(y_{\ell} \mid x_{<\ell}, y_{<\ell}\right) =\prod_{\ell=1}^L p\left(y_{\ell} \mid y_{<\ell}, x_{< \ell}\right),
\end{align*}
where $y_{\ell}$ refers to the $\ell$-th language token, $y_{<\ell}$ to the set of preceding language tokens, and $x_{<\ell}$ to the set of preceding visual tokens. As we focus on modelling the medical literature, here we consider only image-text data~(i.e., no videos).

Following \citet{alayrac_flamingo_2022}, we minimize a joint objective $\mathcal{L}$ over paired and interleaved data:
\begin{align*}
\mathcal{L} = \mathbb{E}_{(x, y) \sim D_{p}}\left[-\sum_{\ell=1}^L \log p\left(y_{\ell} \mid y_{<\ell}, x_{<\ell}\right)\right] + \lambda \cdot \mathbb{E}_{(x, y) \sim D_{i}}\left[-\sum_{\ell=1}^L \log p\left(y_{\ell} \mid y_{<\ell}, x_{<\ell}\right)\right],
\end{align*}
where $D_p$ and $D_i$ stand for the paired and interleaved dataset, respectively. In our case, we use $\lambda=1$.

\subsection{Training}
We performed multi-gpu training on a single node with 8x 80GB NVIDIA A100 GPUs. We trained the model using DeepSpeed ZeRO Stage 2: Optimizer states and gradients are sharded across devices. To further reduce memory load, we employed the 8-bit AdamW optimizer as well as the memory-efficient attention implementation of PyTorch~2.0. Med-Flamingo was initialized at the checkpoint of the Open-Flamingo model and then pre-trained for 2700 steps~(or 6.75 days in wall time, including the validation steps), using 50 gradient accumulation steps and a per-device batch size of 1, resulting in a total batch size of 400. The model has $1.3B$ trainable parameters~(gated cross attention layers and perceiver layers) and roughly $7B$ frozen parameters~(decoder layers and vision encoder), which results in a total of $8.3B$ parameters. Note that this is the same number parameters as in the OpenFlamingo-9B model~(version 1). 

\section{Evaluation}\label{sec:exp-setup}
\subsection{Automatic Evaluation}
\paragraph{Baselines}
To compare generative VQA abilities against the literature, we consider different variants of the following baselines:
\begin{enumerate}
    \item MedVINT~\cite{zhang2023pmc}, a visual instruction-tuned VLM based on Llama. As this model was not designed to do few-shot learning~(e.g. the image information is prepended to the overall input), we report two modes for MedVINT: zero-shot and fine-tuned, where the model was fine-tuned on the training split of the VQA dataset. Since the rather small Visual-USMLE dataset has no separate training split, we ommit the fine-tuned baseline for that dataset. We used the MedVInT-TD model with PMC-LLaMA and PMC-CLIP backbones.
   
    \item OpenFlamingo~\cite{awadalla2023}, a powerful VLM which was trained on general-domain data, and which served as the base model to train Med-Flamingo. We report both zero-shot and few-shot performance. We expect Flamingo-type models to shine in the few-shot setting which they are designed for~(as already the pre-training task includes multiple interleaved image-text examples). 
\end{enumerate}

\paragraph{Evaluation datasets}
To evaluate our model and compare it against the baselines, we leverage two existing VQA datasets from the medical domain~(VQA-RAD and PathVQA). Upon closer inspection of the VQA-RAD dataset, we identified severe data leakage in the official train / test splits, which is problematic given that many recent VLMs fine-tune on the train split. To address this, we created a custom train / test split by seperately splitting images and questions~(each 90\% / 10\%) to ensure that no image or question of the train split leaks into the test split. On these datasets, $6$ shots were used for few-shot.

Furthermore, we create Visual USMLE, a challenging multimodal problem set of $618$ USMLE-style questions which are not only augmented with images but also with a case vignette and potentially tables of laboratory measurements. The Visual USMLE dataset was created by adapting problems from the Amboss platform~(using licenced user access). To make the Visual USMLE problems more actionable and useful, we rephrased the problems to be open-ended instead of multiple-choice. This makes the benchmark harder and more realistic, as the models have to come up with differential diagnoses and potential procedures completely on their own---as opposed to selecting the most reasonable answer choice from few choices. Figure~\ref{sfig:vusmle} gives an overview of the broad range of specialties that are covered in the dataset, greatly extending existing medical VQA datasets which are narrowly focused on radiology and pathology. For this comparatively small dataset, instead of creating a training split for finetuning, we created a small train split of $10$ problems which can be used for few-shot prompting.
For this dataset~(with considerably longer problems and answers), we used only $4$ shots to fit in the context window.

\paragraph{Evaluation metrics}
Previous works in medical vision-language modelling typically focused scoring all available answers of a VQA dataset to arrive at a classification accuracy. However, since we are interested in \emph{generative} VQA~(as opposed to post-hoc scoring different potential answers), for sake of clinical utility, we employ the following evaluation metrics that directly assess the quality of the generated answer:
\begin{enumerate}
    \item Clinical evaluation score, as rated by three medical doctors~(including one board-certified radiologist) using a human evaluation app that we developed for this study. More details are provided in Section~\ref{sec:human_eval}.
    \item BERT similarity score~(BERT-sim), the F1 BERT score between the generated answer and the correct answer~\cite{bert-score}.
    \item Exact-match, the fraction of generated answers that exactly match~(modulo punctuation) the correct answer. This metric is rather noisy and conservative as useful answers may not lexically match the correct answer.
\end{enumerate}

\subsection{Human evaluation} \label{sec:human_eval}
\begin{figure*}[h]
\centering
\includegraphics[width=0.9\linewidth]{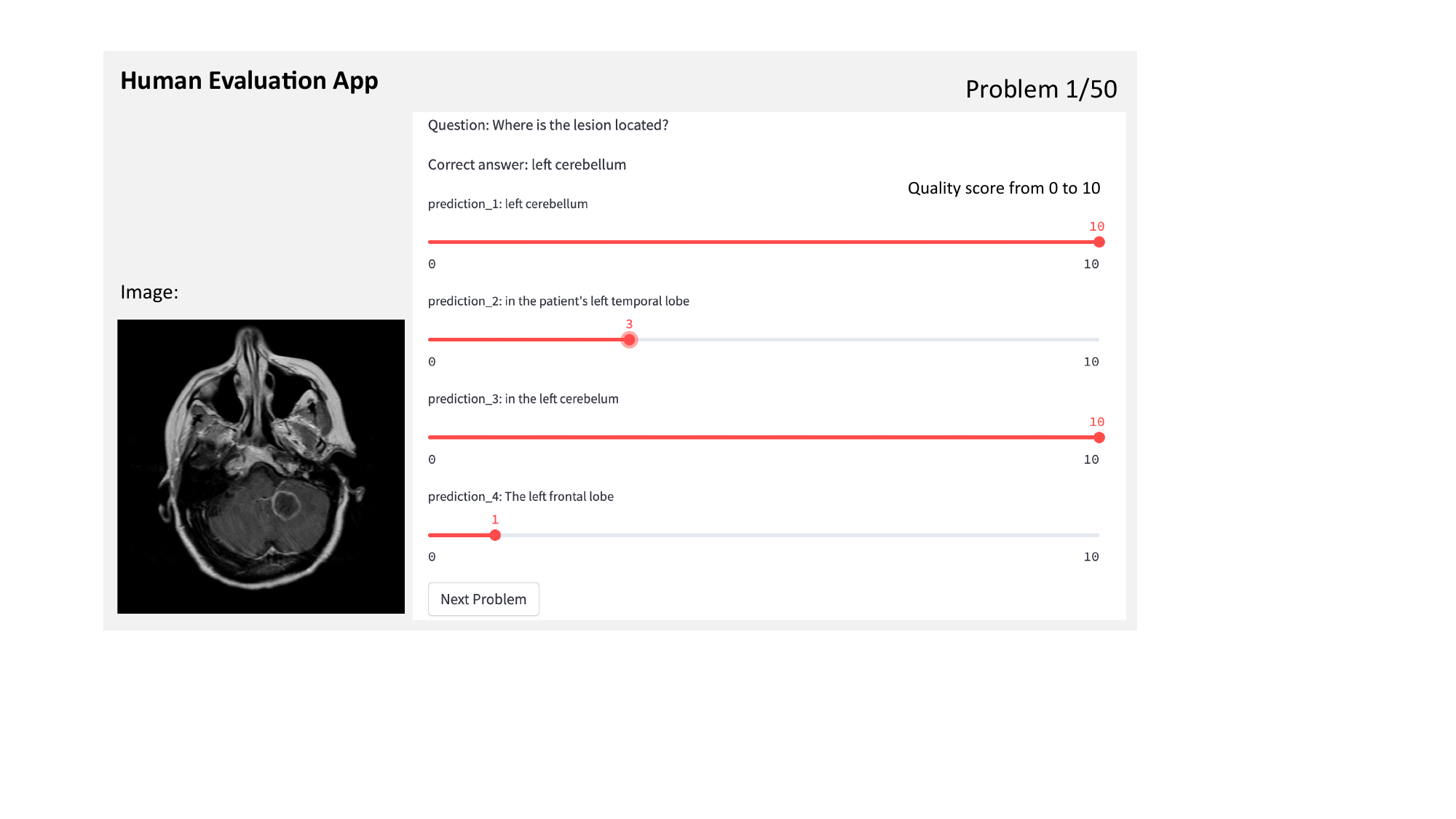}
\caption{Illustration of our Human evaluation app that we created for clinical experts to evaluate generated answers.}
\label{fig:app}
\end{figure*}

We implemented a human evaluation app using Streamlit to visually display the generative VQA problems for clinical experts to rate the quality of the generated answers with scores from $0$ to $10$.  Figure~\ref{fig:app} shows an examplary view of the app. For each VQA problem, the raters are provided with the image, the question, the correct answer, and a set of blinded generations~(e.g., appearing as "prediction\_1" in Figure~\ref{fig:app}), that appear in randomized order.

\subsection{Deduplication and leakage}
During the evaluation of the Med-Flamingo model, we were concerned that there may be leakage between the pre-training datasets~(PMC-OA and MTB) and the down-stream VQA datasets used for evaluation; this could inflate judgements of model quality, as the model could memorize image-question-answer triples. 

To alleviate this concern, we performed data deduplication based upon pairwise similarity between images from our pre-training datasets and the images from our evaluation benchmarks. To detect similar images, in spite of perturbations due to cropping, color shifts, size, etc, we embedded the images using Google's Vision Transformer, preserving the last hidden state as the resultant embedding ~\cite{dosovitskiy2020vit}. We then found the k-nearest neighbors to each evaluation image from amongst the pre-training images (using the FAISS library) ~\cite{johnson2019billion}. We then sorted and visualized image-image pairs by least euclidean distance; we found that images might be duplicates until a pairwise distance of around 80; beyond this point, there were no duplicates. 

This process revealed that the pretraining datasets leaked into the PVQA evaluation benchmark. Out of 6700 total images in PVQA test set, we judged 194 to be highly similar to images in the pretraining datasets, and thus, we removed them from our down-stream evaluation.

\section{Results}

In our experiments, we focus on generative medical visual question answering~(VQA). While recent medical VLMs predominantly performed VQA in a non-generative but rather discriminative manner~(i.e., by scoring different answer choices), we believe that this ex-post classification to carry less clinical usefulness, than directly generating responses. 
On the other hand, generative VQA is more challenging to evaluate, as automated metrics suffer from significant limitations as they do not fully capture the domain-specific context. Thus, we perform a human evaluation study where clinical experts review model generations~(blinded) and score them~(between 0 and 10) in terms of clinical usefulness.

\paragraph{Conventional VQA datasets}
Table~\ref{tab:vqarad} shows the results for VQA-RAD, the radiological VQA dataset for which we created custom splits to address leakage~(see Section\ref{sec:exp-setup}). Med-Flamingo few-shot shows strong results, improving the clinical eval score by $\sim 20\%$ over the best baseline. In this dataset, the auxiliary metrics are rather aligned with clinical preference. Finetuning the MedVINT baseline did not lead to improved performance on this dataset which may be due to its small size. MedVINT zero-shot outperforms the other zero-shot ablations which may be partially attributed to its instruction tuning step on PMC-VQA.

\begin{table*}[b]
\centering
   \begin{tabular}{lrrr}
\toprule
{VQA-RAD} & Clinical eval. score &  BERT-sim &  Exact-match \\
\midrule

MedVINT zero-shot  &     4.63    &  0.628 &     0.167 \\
MedVINT fine-tuned ($\sim2K$ samples)  &       2.87    &  0.611 &     0.133 \\
OpenFlamingo zero-shot  &   4.39   &  0.490 &     0.000 \\
OpenFlamingo few-shot  &    \underline{4.69}   &  \underline{0.645} &     \textbf{0.200} \\
Med-Flamingo zero-shot &   3.82    &  0.480 &     0.000 \\
Med-Flamingo few-shot   &   \textbf{5.61}   &  \textbf{0.650} &     \textbf{0.200} \\
\bottomrule
\end{tabular}
\caption{Performance metrics on the VQA-Rad dataset. Best scores are shown in bold. We put emphasis on the clinical evaluation score. BERT-sim may not fully capture the fine-grained medical details. Exact-match is quite noisy and brittle, but conservative. The fine-tuned baseline did not improve over zero-shot which could be explained by the small dataset size in combination with our custom splits which were created to prevent leakage.}
\label{tab:vqarad}
\end{table*}

\begin{table*}[tbp]
    \centering
    \begin{tabular}{lccc}
    \toprule
    {Path-VQA} & Clinical eval. score &  BERT-sim &  Exact-match \\
    \midrule
    MedVINT zero-shot      & 0.13 &  0.608 & 0.272 \\
    MedVINT fine-tuned~($\sim20K$ samples)            & 1.23 &  \textbf{0.723} & \textbf{0.385} \\
    OpenFlamingo zero-shot & \textbf{2.16} &  0.474 & 0.009 \\
    OpenFlamingo few-shot  & \underline{2.08} &  0.669 & 0.288 \\
    Med-Flamingo zero-shot & 1.72 &  0.521 & 0.120 \\
    Med-Flamingo few-shot  & 1.81 &  \underline{0.678} & \underline{0.303} \\
    \bottomrule
    \end{tabular}
    \caption{Performance metrics on the PathVQA dataset. Best scores are shown in bold. Across models, this dataset showed lowest clinical performance among all evaluation datasets. This highlights a performance deficit in pathology across models, and demonstrates that previous classification-based metrics severely overestimated the performance of general medical VLMs in this specialty.}
    \label{tab:pvqa}
\end{table*}

\begin{figure*}[tbp]
\centering
\includegraphics[width=0.9\linewidth]{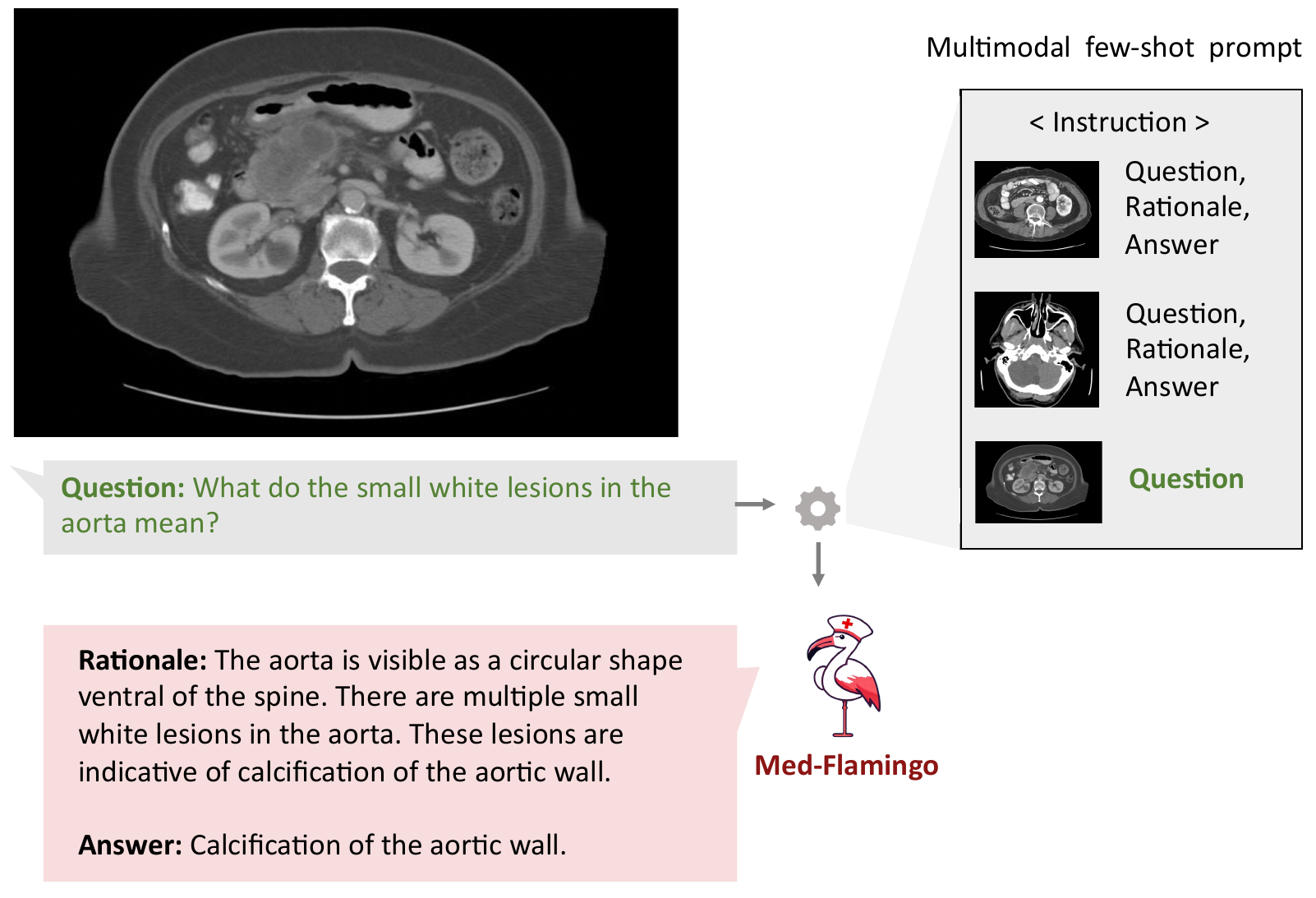}
\caption{Multimodal medical few-shot prompting illustrated with an example. Few-shot prompting here allows users to customize the response format, \emph{e.g.}, to provide rationales for the provided answers. In addition, multimodal few-shot prompts potentially offer the ability to include relevant context retrieved from the medical literature.}
\label{fig:rationale}
\end{figure*}

\begin{table*}[t]
    \centering
    \begin{tabular}{lcc}
    \toprule
    {Visual USMLE} & Clinical eval. score &  BERT-sim  \\
    \midrule
    MedVINT zero-shot      & 0.41 &  0.421  \\
    OpenFlamingo zero-shot & \underline{4.31} &  \textbf{0.512}   \\
    OpenFlamingo few-shot  & 3.39 &  0.470 \\
    Med-Flamingo zero-shot & 4.18 &  \underline{0.473}  \\
    Med-Flamingo few-shot  & \textbf{4.33} &  0.431  \\
    \bottomrule
    \end{tabular}
    \caption{Performance metrics on the Visual USMLE dataset. Best scores are shown in bold. Due to rather lenghty correct answers, the Exact-match metric was not informative as it was constantly $0$ on this dataset.}
    \label{tab:vusmle}
\end{table*}
Table~\ref{tab:pvqa} shows for the results for Path-VQA, the pathology VQA dataset. Compared to the other datasets, all models overall perform poorer on the Path-VQA dataset in terms of clinical evaluation score. We hypothesize that this has to do with the fact the models are not pre-trained on actual large-scale and fine-grained pathology image datasets, but only on a rather small amount of pathology literature~(which may not be enough to achieve strong performance). For instance, Figure~\ref{fig:books} shows that only a small fraction of our training data covers pathology. In the automated metrics~(BERT-sim and exact-match), Med-Flamingo improves upon the OpenFlamingo baseline, however the overall quality does not improve~(as seen in the clinical evaluation score). MedVINT was fine-tuned on a sizeable training split which results in strong automated metrics, but did not result in a clinical evaluation score that matches any Flamingo variant.

\paragraph{Visual USMLE}
Table~\ref{tab:vusmle} shows the results for the Visual USMLE dataset. Med-Flamingo~(few-shot) results in the clinically most preferrable generations, whereas OpenFlamingo~(zero-shot) is a close runner-up. As the ground truth answers were rather lengthy paragraphs, exact match was not an informative metric~(constant 0 for all methods). The few-shot prompted models lead to lower automated scores than their zero-shot counterparts, which we hypothesize has to do with the fact that the USMLE problems are long~(long vignettes as well as long answers) which forced us to summarize the questions and answers when designing few-shot prompts (for which we used GPT-4). Hence, it's possible that those prompts lead to short answers that in terms of BERT-sim score may differ more from the correct answer than a more wordy zero-shot generation.

\paragraph{Across datasets}
Overall, we find that Med-Flamingo's multimodal in-domain few-shot learning abilities lead to favorable generative VQA performance, leading to the lowest average rank of $1.67$ in terms of clinical evaluation score as averaged across all evaluation datasets. As runner-up, OpenFlamingo zero-shot achieves a rank of $2.33$. 

\begin{figure*}[tbp]
\centering
\includegraphics[width=0.9\linewidth]{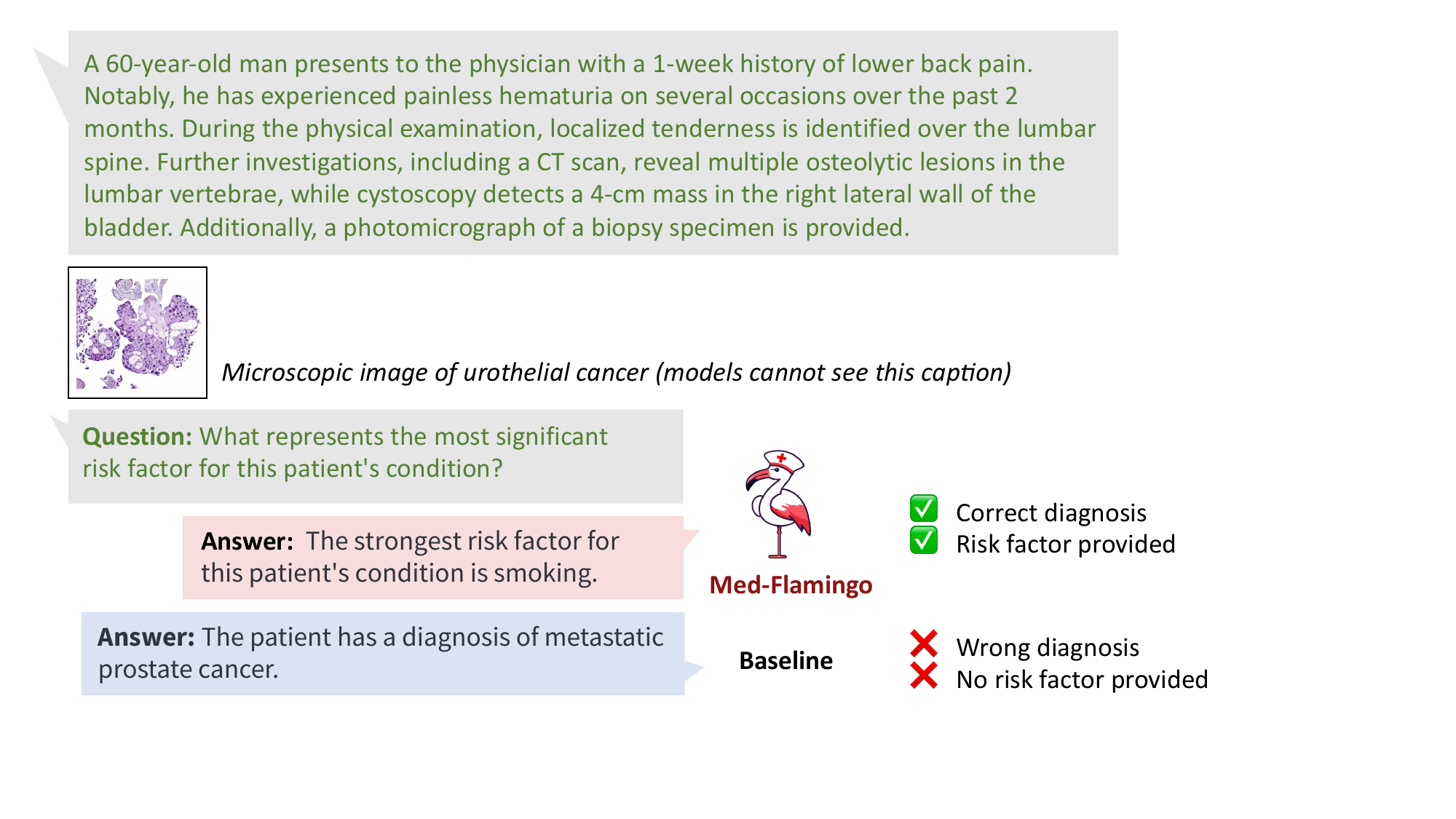}
\caption{Example of a Visual USMLE problem.}
\label{fig:ex1}
\end{figure*}

\paragraph{Qualitative analysis}
Finally, we showcase few examples of Med-Flamingo generations in more detail in Figures~\ref{fig:teaser},\ref{fig:rationale}, and \ref{fig:ex1}. Figure~\ref{fig:rationale} exemplifies that a medical few-shot learner like Med-Flamingo can be prompted to generate rationale for its VQA answer. The shown example is impressive in that the rationale is visually guiding the reader towards the object of interest~(calcification of the aortic wall). We note, however, that at this stage, few-shot multimodal prompted rationales may not be robust, especially when a model arrives at a wrong answer.

Figures~\ref{fig:teaser} and \ref{fig:ex1} showcase two example problems from the Visual USMLE dataset. The problem descriptions were slightly rephrased and summarized using GPT-4 for display. In Figure~\ref{fig:ex1}, Med-Flamingo generates the correct answer while not mentioning the underlying diagnosis~(urothelial cancer) as it was not asked for. By contrast, we observed baselines to directly diagnose the patient (instead of answering the actual question in a targeted way). The problem in Figure~\ref{fig:teaser} illustrates that Med-Flamingo has the ability to integrate complex medical history information together with visual information to synthesize a comprehensive diagnosis that draws from the information of both modalities.

\section{Discussion}
In this paper, we presented Med-Flamingo, the first medically adapted multimodal few-shot learner. While this is an early proof-of-concept for a medical multimodal few-shot learner, we expect to see significant improvements with increased model and data scale, more thoroughly cleaned data, as well as with alignment to human preference via instruction tuning or explicit optimization for preferences.

We expect that the rise of multimodal medical few-shot learners will lead to exciting opportunities with regard to model explainability~(via rationale generation) as well as grounding the model in verified sources~(via multimodal retrieval to augment the few-shot prompt). Thereby, our work serves as a first step towards more generalist medical AI models~\cite{moor2023foundation}.

\paragraph{Limitations} This work demonstrates a proof-of-concept. As such, Med-Flamingo is \emph{not} intended nor safe for clinical use. In all VLMs we analyzed, hallucinations were observed. Furthermore, as Med-Flamingo is a pre-trained model without further instruction or preference tuning, it is possible that the model occasionally outputs low-quality generations. 

\paragraph{Future work} It will be an exciting route for future work to further train Med-Flamingo on clinical data, high-resolution medical image datasets as well as 3D volumes and medical videos. While current general-purpose medical VLMs are pre-trained on the broad medical literature~(\emph{i.e.,} they are only ``book-smart"), also learning from diverse patient data directly will become crucial for down-stream applications.

\section*{Acknowledgments}
We thank Rok Sosič for his technical support in the data preprocessing.

\bibliography{main}
\bibliographystyle{iclr2023_conference}

\newpage

\appendix
\section{Appendix}

\subsection{Additional details for MTB dataset}

\begin{figure}[h]
\centering
\includegraphics[width=1\linewidth]{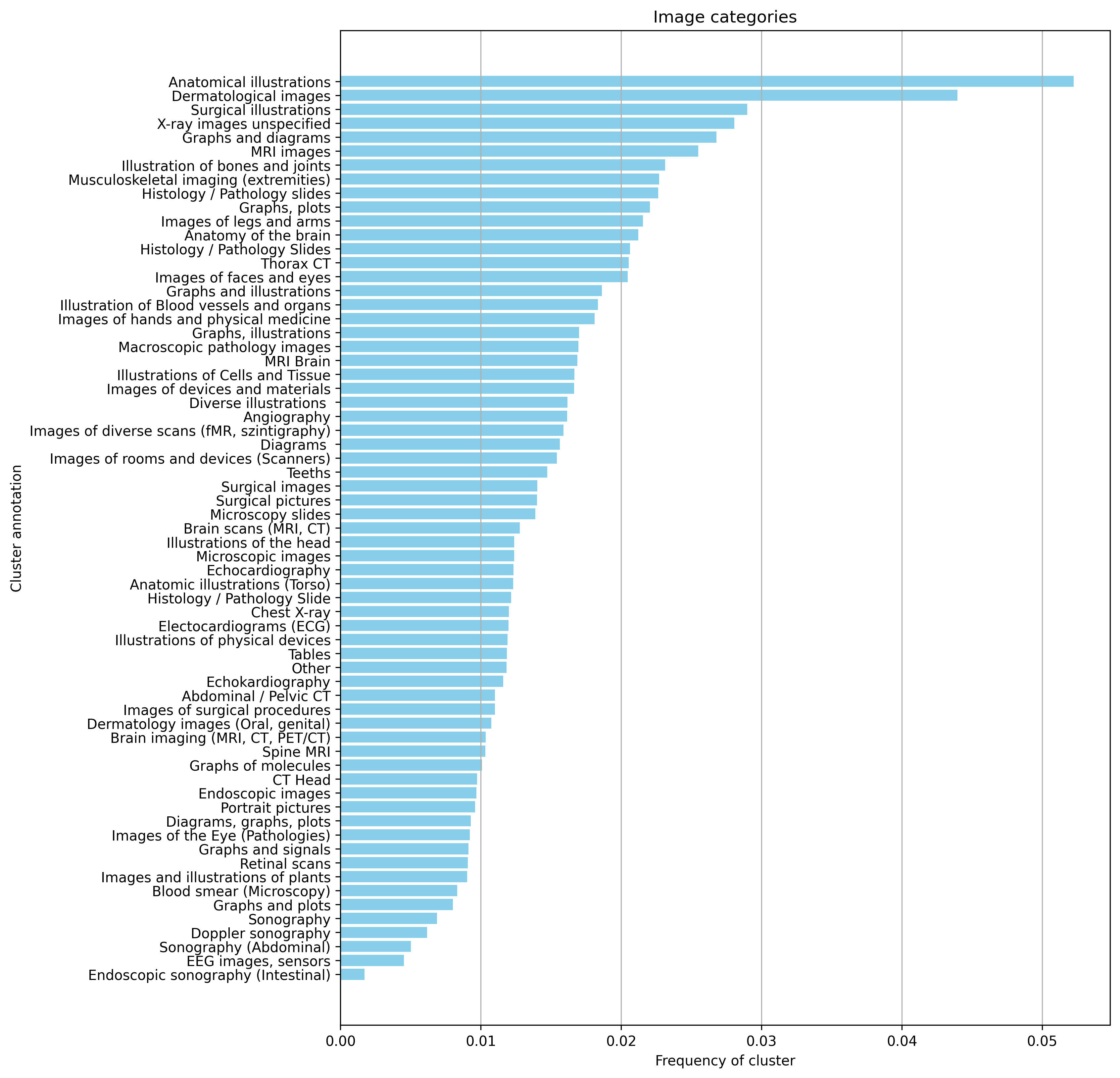}
\caption{Distribution of manually annotated image clusters in the MTB dataset.}
\label{sfig:barplot}
\end{figure}

\paragraph{Clustering the images} In a post-hoc analysis, we clustered the image embeddings of the MTB dataset into a large number of clusters~(100) and manually reviewed examples of each cluster to assign an annotation. We discard noisy or unclear clusters and display the remaining clusters and their frequency in Figure~\ref{sfig:barplot}.

\paragraph{Classification of book titles}
Here, we provide further details about the creation of Figure~\ref{fig:books}. Table~\ref{tab:categories} lists the categories used to prompt the Claude-1 model to classify each book title. We initially prompted with 3 more very rare categories~(Geriatrics, Occupational medicine, Space medicine), but merge them into the "Other" group for visualization purposes.

\begin{table}[ht]
\centering
\begin{tabularx}{\textwidth}{X X X}
Neuroscience/Neurology & Obstetrics and Gynecology & Infectious Diseases \\
Radiology & Dermatology & Family medicine \\
Oncology & Immunology & Biomedical engineering \\
Surgery & Dentistry / Orthodontics & Anesthesiology \\
Cardiology & Ophthalmology & Physiology \\
Psychiatry & Pediatrics & Medical history \\
Pharmacology & Pathology & Nursing \\
Herbal medicine & Anatomy & Otolaryngology \\
Orthopedics & Gastroenterology & Hematology \\
Nutrition & Endocrinology & Urology \\
Internal Medicine & Genetics & Pulmonology \\
Sports Medicine & Medical Research and Statistics & Emergency Medicine \\
Cell Biology and Histology & Pain medicine & Public Health and Epidemiology \\
Forensics & Biochemistry & Nephrology \\
Critical care medicine & Medical Ethics & Veterinary medicine \\
Physical Medicine and Rehabilitation & Health informatics & Mindfulness \\
 Other && \\
\end{tabularx}

\caption{List of 49 Categories (and "Other") used for visualing the MTB dataset in Figure~\ref{fig:books}}
\label{tab:categories}
\end{table}

\subsection{Additional details for Visual USMLE dataset}

\begin{figure}[h]
\centering
\includegraphics[width=1\linewidth]{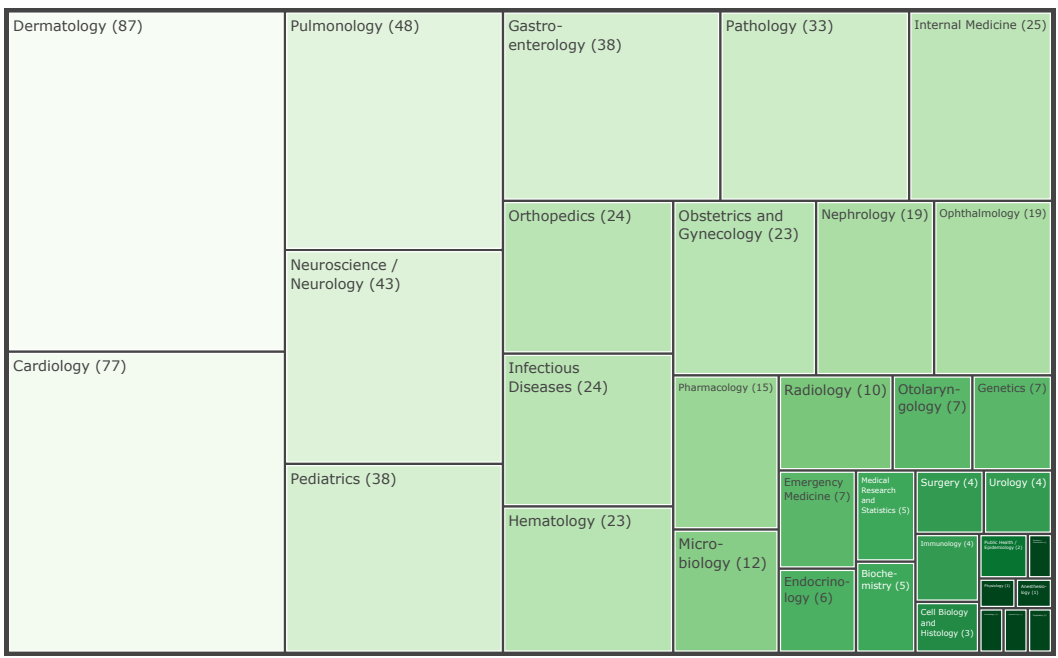}
\caption{Distribution of specialty topics in the Visual USMLE dataset, as classified by Claude-1 using the categories provided in Table~\ref{tab:categories}.}
\label{sfig:vusmle}
\end{figure}

\end{document}